\pdfoutput=1

\documentclass[11pt]{article}

\usepackage[]{emnlp2021}

\usepackage{times}
\usepackage{latexsym}

\usepackage[T1]{fontenc}

\usepackage[utf8]{inputenc}

\usepackage{microtype}
\usepackage[hyperref]{}
\usepackage{times}
\usepackage{latexsym}
\usepackage{graphicx}

\usepackage{microtype}
\usepackage{adjustbox}

\title{Rethinking Crowd Sourcing for Semantic Similarity\thanks{The authors are grateful to Yandex.Toloka for the support of the research project.}}

\author{Shaul Solomon\\
Y-Data\\
Tel Aviv, Israel \\
\texttt{shaulsolomon@gmail.com} \\
\And
 Adam Cohn\\
Y-Data\\
Tel Aviv, Israel \\
\And Hernan Rosenblum\\
Y-Data\\
Tel Aviv, Israel \\
\AND
 Chezi Hershkovitz\\
Y-Data\\
Tel Aviv, Israel \\ 
\And
Ivan P. Yamshchikov\\
 LEYA Laboratory\\
 Yandex, Higher School of Economics\\
St. Petersburg, Russia
}

\date{}

\begin{document}
\maketitle
\begin{abstract}
Estimation of semantic similarity is crucial for a variety of natural language processing (NLP) tasks. In the absence of a general theory of semantic information, many papers rely on human annotators as the source of ground truth for semantic similarity estimation. This paper investigates the ambiguities inherent in crowd-sourced semantic labeling. It shows that annotators that treat semantic similarity as a binary category (two sentences are either similar or not similar and there is no middle ground) play the most important role in the labeling. The paper offers heuristics to filter out unreliable annotators and stimulates further discussions on human perception of semantic similarity.
\end{abstract}

\section{Introduction}

Human-labeled datasets are routinely used as golden datasets for benchmarking NLP algorithms. For some NLP tasks like Part-of-Speech Tagging or Named Entity Recognition the labeling criteria are formulated rigorously, for others rigorous formulation is lacking. A lot of baselines in modern NLP rely on the idea that certain aspects of natural language are understood intuitively by human annotators. This implicit assumption is often the only argument for some form of ontological consistency of obtained evaluations on a given dataset - i.e. that semantics are definitive and unambiguous. This paper demonstrates that this assumption does not hold for semantic similarity measures. It also finds that using domain-specific features of the labeling process, one could detect unreliable annotators and thus significantly affect the results of the labeling. The contributions of this paper are as follows:
\begin{itemize}
\item it highlights that intuitive understanding of semantic similarity varies across language speakers. In the absence of a universal unsupervised semantic similarity measure, these differences lead to the implicit noise in the research outcomes; 
\item using human assessment on thirty-five thousand labeled pairs of sentences the paper explores various inconsistencies present in the labeling; 
\item it proposes five possible heuristics to filter unreliable annotators and evaluates the impact of such filtering on various unsupervised semantic similarity measures.
\end{itemize}

\section{Related Work}

Large scale annotated datasets (Treebank, Imagenet, and many others) have shown to dramatically increase success in many sub-fields of Machine Learning. However, they are extremely expensive and take a long time to develop. One well-established method for obtaining large-scale labeled datasets is crowd-sourcing. In recent years we have seen the rise of various crowd-sourcing services, which provide non-expert annotated labels. By outsourcing the labeling process to external services it's possible to scale horizontally, but researchers face significant challenges to both ensure the quality of the labels and validate that the data is labeled according to the criteria of the task. When dealing with non-expert annotators, the issue of quality assurance arises. Most requesters rely on redundancy (Majority Vote) or use some form of Golden Dataset to filter out unreliable annotators. Beyond the classical methods, many statistical methods have been proposed to address the issue.

Dawid \& Skene \citep{Dawid1979MaximumLE} initially proposed an Expectation-Maximization algorithm to predict the error rate for each annotator. Many other probabilistic models have been proposed \citep{Whitehill2009WhoseVS, Raykar2010LearningFC} to approximate both annotator error and bias \citep{Ipeirotis2010QualityMO}, and the difficulty of particular labels and models \citep{Sheng2008GetAL}. While most of these models have been intended to be generalizable, this paper makes the argument that progress in Natural Language Understanding (NLU) requires attention to the domain-specific attributes of semantic data.

\section{Domain-specific annotator attributes for Natural Language}

In any form of communication or usage of language, there are always two necessary elements: the form and the meaning. When the meaning is tightly bound to the form, one can take the form context-free and be able to extract the meaning directly. However, in natural language communication, it would be impossible to parse the intended meaning without some external knowledge base. The so-called \textit{symbol grounding problem} \cite{Harnad1999} states that one can not derive the meaning of a sentence from the syntax alone. Meaning is derived from many sources, the context, the tone of voice, the relationship between interlocutors, etc.

While the NLP community has made great strides developing a better ability to understand the syntactical distribution of a language, we have yet to make any clear headway in modeling meaning \citep{Bender2020ClimbingTN}. Although annotators may internally feel that they have an intuitive sense of semantic preservation, there does not seem to be a consistent agreement between people (and even for the same person in varying circumstances). 

There are several basic challenges that cause such inconsistencies. Firts of all, the overlap between the form and semantics is very fuzzy, \citep{tikhonov2018wrong, tikhonov2019style}; for example, given a pair of sentences in which the only distinction is the sentiment (ex: "I love pizza." vs. "I hate pizza.") human annotators agree that semantics similarity is low while various NLP researchers treat sentiment as style attribute and evaluate these two sentences as semantically similar. Second, there are many possible axes upon which to calculate semantic similarity (communicative intent \cite{westera2019don}, topic identification \cite{peinelt2020tbert}, emotion recognition \cite{franzoni2017semo}); it is not clear how these axes rank when we are after a general measure of semantic similarity. Finally, personal characteristics of the annotator such as implicit understanding of the context or varying background experience systematically affect the judgment of the annotators.

\section{The Data}

To see differences in the semantic tendencies of human annotation, we used several standard paraphrase and style transfer datasets alongside a random selection of sentence pairs from each dataset. Similar to \cite{Yamshchikov2020} the random pairs of sentences are used for the baseline of sentences that have no semantic overlap whatsoever. The paraphrase datasets include different versions of English Bibles \cite{carlson2017zero}, English Paralex dataset\footnote{http://knowitall.cs.washington.edu/paralex/}, and English Paraphrase dataset\footnote{http://paraphrase.org}. The style transfer datasets are the dataset of politeness introduced in \cite{rao2018dear} referred further as GYAFC, and Yelp! Reviews\footnote{https://www.yelp.com/dataset} enhanced with human-written reviews with opposite sentiment provided by \cite{tian18}.

Every pair of sentences was labeled by three independent annotators with a score from 1-5 (1 being dissimilar and 5 being identical). To facilitate further research of human-labeling inconsistencies for the tasks of semantic similarity, we make all collected information on the labeling process available\footnote{https://github.com/Hernanros/SOTA/tree/master/data}.

There are three major sources of \textit{noise} affecting this labeling procedure. The first source of noise are \textbf{unreliable annotators}. These are people who don't give thoughtful responses and randomly fill their answers. Such annotators are present in all crowdsourcing tasks, and there are many methods to filter them out, for example, \citep{Olesen2011,Lofi2013JustAA}. The second source of noise is the \textbf{1-5 labeling scheme} itself. On the one hand, the continuous scale from one to five presents the possibility for an annotator to mark a pair with 3 implying that two sentences are neither similar nor dissimilar. On the other hand, one could argue that by definition, the lack of similarity inherently equates to dissimilarity. The final source of noise could be \textbf{certain personal qualities of the annotators}. For example, certain users could be more radical in their judgment and have a preference to give extreme ratings (1,5) while others might be more moderate and give more centrist ratings (2,4), see \citep{Panda2020ACF}.

\section{Experiments}

We conducted experiments to estimate the impact of these sources of noise on evaluations of unsupervised semantic measures. Initially, we labored under the assumption of a minimal consistency requirement for a measure of semantic similarity, i.e. that random sentences be ranked less similar than non-random sentences on average. However, when trying to validate that assumption by analyzing the similarity score distributions relative to the labels for different measures, we discovered numerous examples of low-quality labels. As a result, we strove to formalize the patterns of noise into clear heuristics that can be applied to any dataset using the metadata available on a publicly available crowdsourcing platform. 

It can be argued that the heuristics proposed below are generalizable to any form of human judgement, due to the inherent ambiguity within written language discussed prior. This is a legitimate claim, and as such gaining a clearer picture of the biases and noise in the data becomes even more crucial for NLP tasks that require any form of human quantitative estimates. The heuristics below are by no means an exhaustive list, but rather to be viewed as a sample of the myriad of factors that need to be addressed as we strive towards a more comprehensive formulation of semantic similarity.

\subsection{Filtering Heuristics for Unreliable Annotators}

We experimented with five different heuristics:
\begin{enumerate}
    \item \textbf{Slow Annotators}: those whose mean labeling time is much greater than the average labeling time time\footnote{We denoted the annotator who had mean labeling time greater than 300 seconds as a slow annotator. This places them in the 98th percentile in terms of average labeling duration in our dataset.}. 
    \item \textbf{Low Variance}: if the variance for all of the labels given by one annotator is lower than 1\footnote{This means that the vast majority of the labels are annotated with the same label by this annotator}.
    \item \textbf{High Random}: remove labelers whose mean semantic similarity score of all random pairs is higher than their mean semantic similarity score for non-random pairs. Among reliable annotators, the random pairs have to score lower than the ones that are semantically similar.
    \item \textbf{Disagreeable Annotators}: using reduced labeling (Scores below 3 collapse into -1, 3 becomes 0, and anything above 3 collapses into 1) we filter any annotator who happens to disagree with an unanimous decision from the other two annotators more than in half of the cases.
    \item \textbf{Sentimentally Dis-aligned Annotators}: as discussed earlier, the relationship between sentiment and semantics is ambiguous, so we wanted to filter out annotators who used the sentiment to determine semantics in an inconsistent way. \footnote{Taking pairs which have a very high word overlap (BLUE score over 0.8) - indicating nearly identical syntactical content - but with sentiment score differences $\geq$ 1.9 (using huggingface's sentiment-analysis pipeline is bound by [-1,1]), we filter out annotators whose labeling variance on those pairs was greater than 1.}
\end{enumerate}

If the annotator corresponds to one of these categories we pronounce this annotator to be unreliable. To make our experiments clear and reproducible we publish the source code, with specification of all dependencies, including external libraries\footnote{https://github.com/Hernanros/SOTA}. 

\subsection{Correlation with Automated Semantic Similarity Metrics}

To estimate how the labeling noise can interfere with the NLP benchmarks that use semantic similarity measurements we took ten of the most used metrics for content preservation and semantic similarity.   \textbf{Word overlap} is calculated as percentage of words that occur in both texts. \textbf{chrF} \cite{popovic2015chrf} is a character n-gram F-score that measures number of n-grams that coincide in input and output. \textbf{Cosine similarity} is calculated in line with \cite{fu2} either with pre-trained GloVe \cite{pennington2014glove} or  FastText word embeddings \cite{joulin2016fasttext}. \textbf{POS-distance} looks for nouns in the input and output and is calculated as a pairwise distance between the embeddings of the found nouns. \textbf{L2 distance} between Elmo \cite{peters2018deep} embeddings of two sentences. \textbf{WMD} \cite{kusner2015word} defines the distance between two documents as an optimal transport problem between the embedded words. \textbf{BLEU} \cite{papineni2002bleu} is one of the most commmon semantic similarity measures. \textbf{ROUGE} \cite{lin2000automated} compares any text to any other (typically human-generated) summary using a recall-oriented approach and unigrams, with bi-grams, and \cite{lin2004automatic} with the longest co-occurring n-grams in sequence. \textbf{Meteor} \cite{banerjee2005meteor} metric is based on a harmonic mean of unigram precision and recall, with recall, weighted higher than precision and some additional features, such as stemming and synonymy matching. Finally, \textbf{BERT score} proposed in \cite{zhang2019bertscore} is a BERT-based estimator of semantic similarity between two pieces of text.

Table \ref{tab:ex1} shows how the automated semantic similarity metrics correlate with human labels and how they correlate after we filter unreliable annotators defined according to the heuristics above\footnote{See Appendix for the resulting experiments with all five heuristics and relative changes in correlation between human labeling and unsupervised semantic similarity metrics depending on the filtering procedure}. It also clearly demonstrates that relatively straight-forward filtering of human labels could add up to nine percentage points to the results of an automated evaluation. This in itself is disturbing, since such changes in performance are often regarded as an improvement in some NLP tasks. 

\begin{table}[t!]
\centering
\scriptsize{\begin{tabular}{lllll}
\textbf{Metrics} &  \textbf{Baseline} &  \textbf{Filtered } & \textbf{Percentage} \\
                &                   &  \textbf{heuristics} & \textbf{increase} \\
ROUGE-1         &  0.61 &             0.65 &             7.6 \% \\
bleu1           &  0.60 &             0.65 &             7.9 \% \\
ROUGE-l         &  0.60 &             0.65 &             8.0 \% \\
BertScore       &  0.59 &             0.64 &             8.5 \% \\
1-gram\_overlap  &  0.59 &             0.64 &             8.0 \% \\
chrfScore       &  0.58 &             0.63 &             7.3 \% \\
L2\_score        &  0.56 &             0.6 &             6.6 \% \\
ROUGE-2         &  0.53 &             0.58 &             8.4 \% \\
fasttext\_cosine &  0.51 &             0.52 &             2.2 \% \\
WMD             &  0.5 &             0.52 &             4.4 \% \\
glove\_cosine    &  0.45 &             0.48 &             4.6 \% \\
bleu            &  0.41 &             0.45 &             9.0 \% \\
POS Dist score  &  0.35 &             0.38 &             7.2 \% \\
\end{tabular}}
\caption{The correlation between automated semantic similarity metrics and the human labels over all datasets, and the same correlation  when unreliable annotators are filtered out. The automated metrics improve from 2 to 9 percentage points depending on the metric.}
 \label{tab:ex1}
\end{table}

\begin{table}[t!]
\centering
\scriptsize{
\begin{tabular}{lllll}
\textbf{Metric} &  \textbf{Baseline} &  \textbf{Radicals} &  \textbf{Baseline} &  \textbf{Centrists} \\
                &  \textbf{Radicals} &  \textbf{after filter} &  \textbf{Centrists} &  \textbf{after filter} \\
ROUGE-1         &          0.70 &          0.70 &           0.28 &           0.45 \\
bleu1           &          0.69 &          0.69 &           0.28 &           0.46 \\
ROUGE-l         &          0.69 &          0.69 &           0.28 &           0.46 \\
BertScore       &          0.69 &          0.69 &           0.28 &           0.46 \\
1-gram\_overlap  &          0.68 &          0.68 &           0.28 &           0.45 \\
chrfScore       &          0.67 &          0.67 &           0.28 &           0.44 \\
L2\_score        &          0.64 &          0.65 &           0.25 &           0.39 \\
ROUGE-2         &          0.61 &          0.61 &           0.26 &           0.41 \\
fasttext\_cosine &          0.57 &          0.57 &           0.21 &           0.32 \\
WMD             &          0.57 &          0.56 &           0.21 &           0.33 \\
glove\_cosine    &          0.51 &          0.51 &           0.19 &           0.23 \\
bleu            &          0.47 &          0.48 &           0.21 &           0.32 \\
POS Dist score  &          0.41 &          0.40 &           0.16 &           0.28 \\
\end{tabular}}
\caption{The Baseline correlation without filtering and the improvement after filtering unreliable annotators for Radical and Centrist Annotators independently.}
 \label{tab:ex2}
\end{table}

In all our experiments situations we see that the combination of \textbf{Low Variance} and \textbf{High Random} filtering heuristics has the strongest impact on the correlation with automated evaluation methods. Under certain circumstances, heuristics based on \textbf{Disagreeable Labelers} and \textbf{Sentimentally Dis-aligned Labelers} also increase the correlation. On the other hand, filtering \textbf{Slow Annotators} out only hurts the performance. 

Table \ref{tab:ex2} shows a more nuanced picture of correlations between the automated metrics and the labels of different annotators. We denote those annotators who selected \{1,5\} over 50\% of the time as \textit{Radical} and those who selected \{2,4\} as \textit{Centrist}; the label 3 was ignored in calculations of these criteria, and only annotators with variance above 1 were included. Comparing results in Table \ref{tab:ex1} and Table \ref{tab:ex2} one could see that radical annotators play a major part in the resulting overall correlations between human labels and automated semantic similarity metrics. Moreover, filtering unreliable annotators only affects correlations of the labels given by Centrists. This either shows that treating semantic similarity as a binary value when using crowd-sourced human labels might be beneficial for less ambiguous results or hints that current unsupervised metrics of semantic similarity have hard time capturing nuance that some human annotators see.

\section{Conclusion}

This paper is an attempt to quantify the inherent ambiguity prevalent in any NLP task that relies on human judgment as a measure of semantic similarity. It demonstrates that a simple heuristic curation of human annotation could give up to \textbf{9 extra percentage points} in terms of the model performance estimated with some unsupervised semantic similarity measure. 

The series of experiments conducted in the paper provides several rules of thumb to reduce the ambiguity of human labels: (1) when labeling treat semantic similarity as a binary feature asking if two texts are similar or not, (2) add sentence pairs where there is no semantic similarity whatsoever and filter unreliable annotators that make mistakes on these pairs, (3) majority vote improves the consistency of your data but it is not as good, as filtering out annotators with label variance and annotators that systematically make mistakes with obviously dissimilar sentence pairs. 

\bibliography{eacl2021}
\bibliographystyle{acl_natbib}

\clearpage

\begin{table*}[t]
\centering
\scriptsize{
\begin{tabular}{llllllll}
base &            \textbf{bleu} &           \textbf{bleu1} &    \textbf{glove cosine} & \textbf{fasttext cosine} &       \textbf{BertScore} &       \textbf{chrfScore} &  \textbf{POS Dist score} \\
correlation  &            0.41 &            0.60 &            0.45 &            0.51 &            0.59 &            0.58 &            0.35     \\
$[1]$              &   0.39 (-4.34\%) &   0.58 (-3.54\%) &   0.44 (-3.01\%) &   0.49 (-3.22\%) &   0.57 (-3.46\%) &   0.56 (-3.56\%) &   0.34 (-4.09\%)\\
$[2]$              &   0.44 (+7.31\%) &   0.65 (+7.47\%) &   0.48 (+5.42\%) &   0.53 (+4.59\%) &   0.64 (+7.56\%) &   0.63 (+7.21\%) &   0.38 (+8.35\%) \\
$[3]$              &   0.43 (+4.09\%) &   0.63 (+4.12\%) &   0.47 (+3.44\%) &   0.53 (+3.04\%) &   0.62 (+4.09\%) &   0.61 (+3.84\%) &   0.36 (+3.88\%)\\
$[4]$              &   0.41 (+0.01\%) &   0.60 (+0.01\%) &   0.45 (+0.04\%) &   0.51 (+0.04\%) &   0.59 (+0.01\%) &   0.58 (+0.00\%) &   0.35 (-0.05\%) \\
$[5] $             &   0.38 (-8.53\%) &   0.54 (-9.77\%) &  0.41 (-10.23\%) &  0.45 (-10.78\%) &   0.54 (-9.37\%) &   0.53 (-9.45\%) &  0.31 (-10.54\%) \\
$[1, 2] $          &   0.43 (+4.76\%) &   0.64 (+5.81\%) &   0.47 (+4.14\%) &   0.52 (+2.90\%) &   0.63 (+6.01\%) &   0.61 (+5.38\%) &   0.37 (+5.67\%)\\
$[1, 3]  $         &   0.41 (+0.38\%) &   0.61 (+1.24\%) &   0.46 (+0.85\%) &   0.51 (+0.30\%) &   0.60 (+1.21\%) &   0.59 (+0.93\%) &   0.35 (+0.33\%) \\
$[1, 4]  $         &   0.39 (-4.33\%) &   0.58 (-3.52\%) &   0.44 (-2.97\%) &   0.49 (-3.17\%) &   0.57 (-3.44\%) &   0.56 (-3.55\%) &   0.34 (-4.14\%) \\
$[1, 5]    $       &  0.36 (-11.50\%) &  0.53 (-12.04\%) &  0.40 (-12.08\%) &  0.44 (-12.88\%) &  0.52 (-11.50\%) &  0.51 (-11.79\%) &  0.30 (-13.84\%) \\
$[2, 3] $          &  0.45 (+10.60\%) &  0.67 (+10.69\%) &   0.49 (+8.65\%) &   0.55 (+7.67\%) &  0.66 (+10.91\%) &  0.64 (+10.34\%) &  0.39 (+11.18\%)\\
$[2, 4] $          &   0.44 (+7.31\%) &   0.65 (+7.49\%) &   0.48 (+5.46\%) &   0.53 (+4.64\%) &   0.64 (+7.59\%) &   0.63 (+7.22\%) &   0.38 (+8.32\%) \\
$[2, 5] $          &   0.44 (+7.50\%) &   0.64 (+5.73\%) &   0.46 (+1.90\%) &   0.51 (+0.16\%) &   0.63 (+6.20\%) &   0.62 (+5.46\%) &   0.37 (+5.79\%)  \\
$[3, 4] $          &   0.43 (+4.09\%) &   0.63 (+4.13\%) &   0.47 (+3.47\%) &   0.53 (+3.08\%) &   0.62 (+4.10\%) &   0.61 (+3.84\%) &   0.36 (+3.83\%) \\
$[3, 5] $          &   0.40 (-2.88\%) &   0.58 (-4.38\%) &   0.43 (-5.89\%) &   0.48 (-6.45\%) &   0.57 (-4.22\%) &   0.56 (-4.51\%) &   0.33 (-6.36\%) \\ 
$[4, 5] $    &   0.38 (-8.50\%) &   0.54 (-9.75\%) &  0.41 (-10.18\%) &  0.45 (-10.72\%) &   0.54 (-9.34\%) &   0.53 (-9.43\%) &  0.31 (-10.57\%)  \\
$[1, 2, 3] $       &   0.45 (+8.89\%) &   0.66 (+9.65\%) &   0.49 (+7.87\%) &   0.54 (+6.48\%) &   0.65 (+9.94\%) &   0.64 (+9.11\%) &   0.38 (+9.18\%)  \\
$[1, 2, 4]  $      &   0.43 (+4.76\%) &   0.64 (+5.83\%) &   0.47 (+4.17\%) &   0.52 (+2.95\%) &   0.63 (+6.04\%) &   0.61 (+5.39\%) &   0.37 (+5.63\%) \\
$[1, 2, 5] $       &   0.44 (+6.95\%) &   0.64 (+5.55\%) &   0.46 (+1.81\%) &   0.51 (-0.43\%) &   0.63 (+6.23\%) &   0.61 (+5.14\%) &   0.37 (+5.22\%)  \\
$[1, 3, 4] $       &   0.41 (+0.37\%) &   0.61 (+1.25\%) &   0.46 (+0.88\%) &   0.51 (+0.34\%) &   0.60 (+1.22\%) &   0.59 (+0.94\%) &   0.35 (+0.28\%) \\
$[1, 3, 5] $       &   0.39 (-5.25\%) &   0.57 (-5.95\%) &   0.42 (-7.20\%) &   0.47 (-8.02\%) &   0.56 (-5.76\%) &   0.55 (-6.18\%) &   0.32 (-8.96\%) \\
$[1, 4, 5]$        &  0.36 (-11.47\%) &  0.53 (-12.01\%) &  0.40 (-12.03\%) &  0.44 (-12.82\%) &  0.52 (-11.47\%) &  0.51 (-11.77\%) &  0.30 (-13.87\%) \\
$[2, 3, 4]$        &  0.45 (+10.60\%) &  0.67 (+10.71\%) &   0.49 (+8.69\%) &   0.55 (+7.72\%) &  0.66 (+10.94\%) &  0.64 (+10.36\%) &  0.39 (+11.15\%)  \\
$[2, 3, 5]$        &   0.45 (+9.31\%) &   0.65 (+7.79\%) &   0.47 (+4.41\%) &   0.52 (+2.51\%) &   0.64 (+8.19\%) &   0.63 (+7.36\%) &   0.38 (+7.42\%)  \\
$[2, 4, 5] $       &   0.44 (+7.51\%) &   0.64 (+5.76\%) &   0.46 (+1.94\%) &   0.51 (+0.22\%) &   0.63 (+6.24\%) &   0.62 (+5.48\%) &   0.37 (+5.76\%) \\
$[3, 4, 5] $       &   0.40 (-2.85\%) &   0.58 (-4.35\%) &   0.43 (-5.85\%) &   0.48 (-6.39\%) &   0.57 (-4.20\%) &   0.56 (-4.49\%) &   0.33 (-6.40\%) \\
$[1, 2, 3, 4]$     &   0.45 (+8.89\%) &   0.66 (+9.66\%) &   0.49 (+7.91\%) &   0.54 (+6.53\%) &   0.65 (+9.97\%) &   0.64 (+9.12\%) &   0.38 (+9.15\%) \\
$[1, 2, 3, 5]$     &   0.45 (+9.03\%) &   0.65 (+7.85\%) &   0.48 (+4.53\%) &   0.52 (+2.10\%) &   0.64 (+8.45\%) &   0.63 (+7.25\%) &   0.38 (+7.18\%)\\
$[1, 2, 4, 5]$     &   0.44 (+6.95\%) &   0.64 (+5.57\%) &   0.46 (+1.86\%) &   0.51 (-0.37\%) &   0.63 (+6.27\%) &   0.61 (+5.16\%) &   0.37 (+5.18\%)  \\
$[1, 3, 4, 5]$     &   0.39 (-5.23\%) &   0.57 (-5.92\%) &   0.42 (-7.15\%) &   0.47 (-7.96\%) &   0.56 (-5.73\%) &   0.55 (-6.16\%) &   0.32 (-9.00\%)  \\
$[2, 3, 4, 5]$     &   0.45 (+9.32\%) &   0.65 (+7.82\%) &   0.47 (+4.45\%) &   0.52 (+2.57\%) &   0.64 (+8.23\%) &   0.63 (+7.38\%) &   0.38 (+7.38\%)  \\
$[1, 2, 3, 4, 5] $ &   0.45 (+9.04\%) &   0.65 (+7.88\%) &   0.48 (+4.58\%) &   0.52 (+2.16\%) &   0.64 (+8.50\%) &   0.63 (+7.27\%) &   0.38 (+7.15\%) \\
\end{tabular}}
\caption{The Baseline correlation and the impact of each of the heuristics for each of the metrics.
[1] — slow annotators, [2] — low variance, [3] — high random, [4] disagreeable annotators, [5] — sentimentally dis-aligned annotators}
 \label{tab:ex3}
\end{table*}

\begin{table*}[t]
\centering
\scriptsize{
\begin{tabular}{lllllll}
{}               &         \textbf{ROUGE-1} &         \textbf{ROUGE-2} &         \textbf{ROUGE-l} &        \textbf{L2 score} &             \textbf{WMD} &  \textbf{1-gram overlap} \\
base correlation &            0.61 &            0.53 &            0.60 &            0.56 &            0.50 &            0.59 \\
$[1]$              &   0.59 (-3.51\%) &   0.51 (-3.87\%) &   0.58 (-3.56\%) &   0.54 (-3.24\%) &   0.48 (-3.20\%) &   0.57 (-3.69\%) \\
$[2] $             &   0.66 (+7.54\%) &   0.57 (+7.25\%) &   0.65 (+7.52\%) &   0.60 (+6.97\%) &   0.53 (+6.49\%) &   0.63 (+7.59\%) \\
$[3] $             &   0.63 (+4.05\%) &   0.55 (+3.91\%) &   0.62 (+4.05\%) &   0.59 (+4.22\%) &   0.52 (+4.05\%)&   0.61 (+4.01\%) \\
$[4] $             &   0.61 (+0.01\%) &   0.53 (+0.00\%) &   0.60 (+0.01\%) &   0.56 (+0.03\%) &   0.50 (-0.02\%) &   0.59 (+0.01\%) \\
$[5]$              &   0.55 (-9.97\%) &   0.48 (-8.85\%) &   0.54 (-9.58\%) &   0.51 (-9.83\%) &  0.45 (-10.66\%) &   0.53 (-9.67\%) \\
$[1, 2]$           &   0.65 (+5.87\%) &   0.56 (+5.24\%) &   0.64 (+5.82\%) &   0.59 (+5.43\%) &   0.52 (+4.72\%) &   0.62 (+5.76\%) \\
$[1, 3] $          &   0.62 (+1.17\%) &   0.53 (+0.65\%) &   0.61 (+1.12\%) &   0.57 (+1.57\%) &   0.51 (+1.40\%)  &   0.59 (+0.96\%) \\
$[1, 4] $          &   0.59 (-3.49\%) &   0.51 (-3.86\%) &   0.58 (-3.54\%) &   0.54 (-3.21\%) &   0.48 (-3.22\%) &   0.57 (-3.68\%) \\
$[1, 5] $          &  0.53 (-12.23\%) &  0.47 (-11.50\%) &  0.53 (-11.86\%) &  0.50 (-11.86\%) &  0.44 (-12.68\%) &  0.52 (-12.07\%) \\
$[2, 3]  $         &  0.68 (+10.76\%) &  0.59 (+10.41\%) &  0.67 (+10.65\%) &  0.62 (+10.19\%) &   0.55 (+9.86\%) &  0.65 (+10.78\%) \\
$[2, 4] $          &   0.66 (+7.56\%) &   0.57 (+7.25\%) &   0.65 (+7.54\%) &   0.60 (+7.01\%) &   0.53 (+6.48\%) &   0.63 (+7.61\%) \\
$[2, 5] $          &   0.64 (+5.44\%) &   0.57 (+6.58\%) &   0.64 (+5.88\%) &   0.59 (+4.29\%) &   0.51 (+2.74\%) &   0.62 (+5.95\%) \\
$[3, 4] $          &   0.63 (+4.06\%) &   0.55 (+3.91\%) &   0.62 (+4.06\%) &   0.59 (+4.25\%) &   0.52 (+4.03\%) &   0.61 (+4.01\%) \\
$[3, 5]$           &   0.58 (-4.72\%) &   0.51 (-3.67\%) &   0.57 (-4.28\%) &   0.54 (-4.51\%) &   0.47 (-5.39\%) &   0.56 (-4.39\%) \\ 
$[4, 5] $          &   0.55 (-9.95\%) &   0.48 (-8.83\%) &   0.54 (-9.55\%) &   0.51 (-9.78\%) &  0.45 (-10.68\%) &   0.53 (-9.65\%) \\
$[1, 2, 3]$        &   0.67 (+9.67\%) &   0.58 (+9.09\%) &   0.66 (+9.54\%) &   0.61 (+9.27\%) &   0.54 (+8.58\%)  &   0.65 (+9.55\%) \\
$[1, 2, 4]$        &   0.65 (+5.89\%) &   0.56 (+5.24\%) &   0.64 (+5.83\%) &   0.59 (+5.47\%) &   0.52 (+4.70\%) &   0.62 (+5.77\%) \\
$[1, 2, 5] $       &   0.64 (+5.22\%) &   0.56 (+6.14\%) &   0.64 (+5.67\%) &   0.59 (+4.06\%) &   0.51 (+2.27\%) &     0.62 (+5.66\%) \\
$[1, 3, 4] $       &   0.62 (+1.19\%) &   0.53 (+0.65\%) &   0.61 (+1.13\%) &   0.57 (+1.59\%) &   0.51 (+1.38\%)  &  0.59 (+0.96\%) \\
$[1, 3, 5]$        &   0.57 (-6.29\%) &   0.50 (-5.72\%) &   0.57 (-5.88\%) &   0.53 (-5.91\%) &   0.46 (-6.84\%) &  0.55 (-6.11\%) \\
$[1, 4, 5]$        &  0.54 (-12.21\%) &  0.47 (-11.48\%) &  0.53 (-11.84\%) &  0.50 (-11.81\%) &  0.44 (-12.69\%) & 0.52 (-12.05\%) \\
$[2, 3, 4]$        &  0.68 (+10.78\%) &  0.59 (+10.41\%) &  0.67 (+10.66\%) &  0.62 (+10.24\%) &   0.55 (+9.85\%)&   0.65 (+10.79\%) \\
$[2, 3, 5]$        &   0.66 (+7.52\%) &   0.58 (+8.55\%) &   0.65 (+7.89\%) &   0.60 (+6.52\%) &   0.52 (+4.77\%)&   0.64 (+7.97\%) \\
$[2, 4, 5]$        &   0.64 (+5.47\%) &   0.57 (+6.59\%) &   0.64 (+5.90\%) &   0.59 (+4.35\%) &   0.51 (+2.73\%) &    0.62 (+5.97\%) \\
$[3, 4, 5] $       &   0.58 (-4.70\%) &   0.51 (-3.65\%) &   0.57 (-4.26\%) &   0.54 (-4.47\%) &   0.47 (-5.41\%)&  0.56 (-4.38\%) \\
$[1, 2, 3, 4]$     &   0.67 (+9.69\%) &   0.58 (+9.09\%) &   0.66 (+9.56\%) &   0.61 (+9.31\%) &   0.54 (+8.57\%) &    0.65 (+9.56\%) \\
$[1, 2, 3, 5]$     &   0.66 (+7.55\%) &   0.58 (+8.38\%) &   0.65 (+7.93\%) &   0.60 (+6.55\%) &   0.52 (+4.46\%)  &   0.64 (+7.93\%) \\
$[1, 2, 4, 5]$     &   0.64 (+5.25\%) &   0.56 (+6.15\%) &   0.64 (+5.69\%) &   0.59 (+4.12\%) &   0.51 (+2.26\%)&   0.62 (+5.69\%) \\
$[1, 3, 4, 5]$     &   0.57 (-6.26\%) &   0.50 (-5.71\%) &   0.57 (-5.85\%) &   0.53 (-5.86\%) &   0.46 (-6.86\%) &   0.55 (-6.09\%) \\
$[2, 3, 4, 5]  $   &   0.66 (+7.55\%) &   0.58 (+8.56\%) &   0.65 (+7.91\%) &   0.60 (+6.58\%) &   0.52 (+4.76\%) &   0.64 (+8.0\%)\\
$[1, 2, 3, 4, 5]$  &   0.66 (+7.58\%) &   0.58 (+8.39\%) &   0.65 (+7.95\%) &   0.60 (+6.60\%) &   0.52 (+4.45\%) &   0.64 (+7.96\%)\\
\end{tabular}}
\caption{The Baseline correlation and the impact of each of the heuristics for each of the metrics.
[1] — slow annotators, [2] — low variance, [3] — high random, [4] disagreeable annotators, [5] — sentimentally dis-aligned annotators}
 \label{tab:ex4}
\end{table*}

\end{document}